\documentclass[conference]{IEEEtran}
\IEEEoverridecommandlockouts

\usepackage{cite}
\usepackage{hyperref}
\usepackage{amsmath,amssymb,amsfonts}
\usepackage{algorithmic}
\usepackage{graphicx}
\usepackage{textcomp}
\usepackage{xcolor}
\def\BibTeX{{\rm B\kern-.05em{\sc i\kern-.025em b}\kern-.08em
    T\kern-.1667em\lower.7ex\hbox{E}\kern-.125emX}}
    
\begin{document}

\title{\Large Towards Tangible Immersion for Cobot Programming-by-Demonstration\\\small{Visual, Tactile and Haptic   Interfaces for Mixed-Reality Cobot Automation in Semiconductor Manufacturing}
\vspace{-4mm}
}
\author{\IEEEauthorblockN{\footnotesize David I. Gonzalez-Aguirre, Javier Felip Leon, Javier Felix-Rendon, Roderico Garcia-Leal and Julio C. Zamora Esquivel}
\IEEEauthorblockA{\footnotesize Intel Labs, Intelligent System Research, Human-Robot Collaboration. Hillsboro OR, USA. \\
\scriptsize\{david.i.gonzalez.aguirre, javier.felip.leon,javier.felix.rendon,roderico.garcia.leal.julio.c.zamora.esquivel\}@intel.com}\vspace{-5mm}}

\maketitle

\begin{abstract}
Collaborative robots are transforming automation by executing and adapting to diverse tasks through high-level commands. This progress is fueled by advanced multimodal AI models and human-robot imitation paradigms, enabling compelling real-world applications. Our research integrates human demonstrations to automate tasks with Cobots using intuitive interfaces for data collection and real-time teleoperation, facilitating interactive robot programming with immersive interfaces. Despite significant advancements in this field, challenges persist in deploying cost-effective, multi-user instrumentation at scale, especially in industries handling fragile objects that require precise tactile and visual perception in ultra-clean environments. We propose using multi-cue human demonstrations to define and structure grasping, placement, transport, and perception primitives, enhancing Cobots' manipulation capabilities without requiring retraining. Our mixed-reality approach enables scalable Cobot programming, addressing limitations in object sizes, users' vantage points, and kinematic registrations of head-mounted displays, cameras, and robots. Our research aims to democratize robot programmability via enhanced mixed-reality human-robot interfaces and online imitation, transforming Cobots into efficient helpers.
\end{abstract}


\section{Introduction}

The field of AI-robotics is rapidly evolving, with collaborative robots emerging as intelligent assistants capable of executing diverse physical tasks using high-level prompts as semantic commands \cite{stepputtis2020language}. This evolution is fueled by novel multimodal AI models and data-driven human-robot imitation paradigms, paving the way for unprecedented automation in real-world deployments \cite{black2024pi_0}. A key enabler aspect of these advancements is the coherent acquisition of human demonstrations to train, fine-tune, and validate these approaches \cite{hu2023robofm}. This requires human-robot interfaces that empower users with intuitive simulation and teleoperation mechanisms, while also managing environmental and operational restrictions. Additionally, these interfaces must be cost-effective and scalable to support complex processes like machine tending in clean rooms for semiconductor inspection and manufacturing. Despite advances in AI-driven robot programmability, challenges persist in instrumentation costs and end-effector materials, especially for handling small, fragile objects in diverse batches requiring precise tactile sensing and visual perception. 

\begin{figure*}
    \centering
    \includegraphics[width=1.00\linewidth]{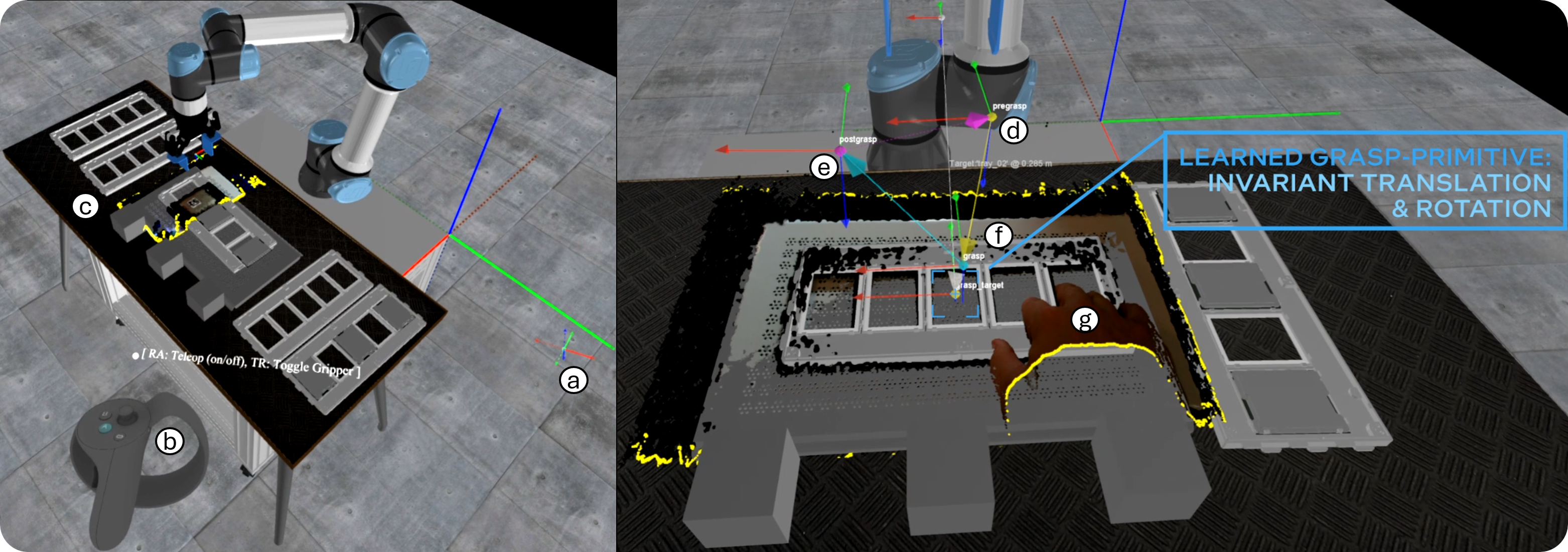}
    \vspace{-6mm}
    \caption{\textbf{Mixed-Reality for Cobot Programming-by-Demonstration.} The real-time visualization of 3D assets within a unified spatio-temporal  kinematic frame enables the overlay of reconstructed surfaces from RGBD data, creating a mixed-reality environment for Cobot teleoperation. This visualization is ideal for developing Cobot programs through annotated task co-execution. a) The kinematic registration reference frame of the head-mounted display, obtained via a simple 3-point annotation, is depicted within the 3D scene. b) Handheld controls facilitate visualization of feasible actions with contextual menus during grasp and placement demonstrations. c) The 3D surface reconstruction combining depth and color streams blends reality with registered CAD/CAM models, showing kinematic frames for grasp primitives in relation to the tray, labeled as d) \textit{pre-grasp}, e) \textit{grasp} and f) \textit{post-grasp}. The 6D-invariant grasp primitive offers high-level composability and low-level robust execution. g) Finally, the user's hand in mixed reality demonstrates the system's low latency, \href{https://www.dropbox.com/scl/fi/gpdgsa8t44gvkwzsdgi3d/Human-Robot-Collaboration_PbD_2023_75MB.mp4?rlkey=gsolb8y6hjsuw57qkyg48xwl6&st=4zgen5ll&dl=0}{see video}.}
    \label{fig:mixed_reality}
    \vspace{-5mm}
\end{figure*}

Our research utilizes multi-cue human demonstrations to define grasping primitives, combining contact points and stable closures within suitable pressure ranges. Mixed reality interfaces enable users to define and execute complex grasp configurations, overcoming sub-millimeter limitations and occluded perspectives. This approach is promising for industries where product size, shape, and materials pose challenges, such as semiconductor, pharmaceutical, and chemical sectors. Our programming-by-demonstration involves real-time human-robot task co-execution, mixed-reality teleoperation, and novel visuo-haptic feedback interfaces, structuring actions into modular robot workflows, see Fig.~\ref{fig:mixed_reality}.
\begin{figure}
    \centering
\includegraphics[width=1.0\linewidth]{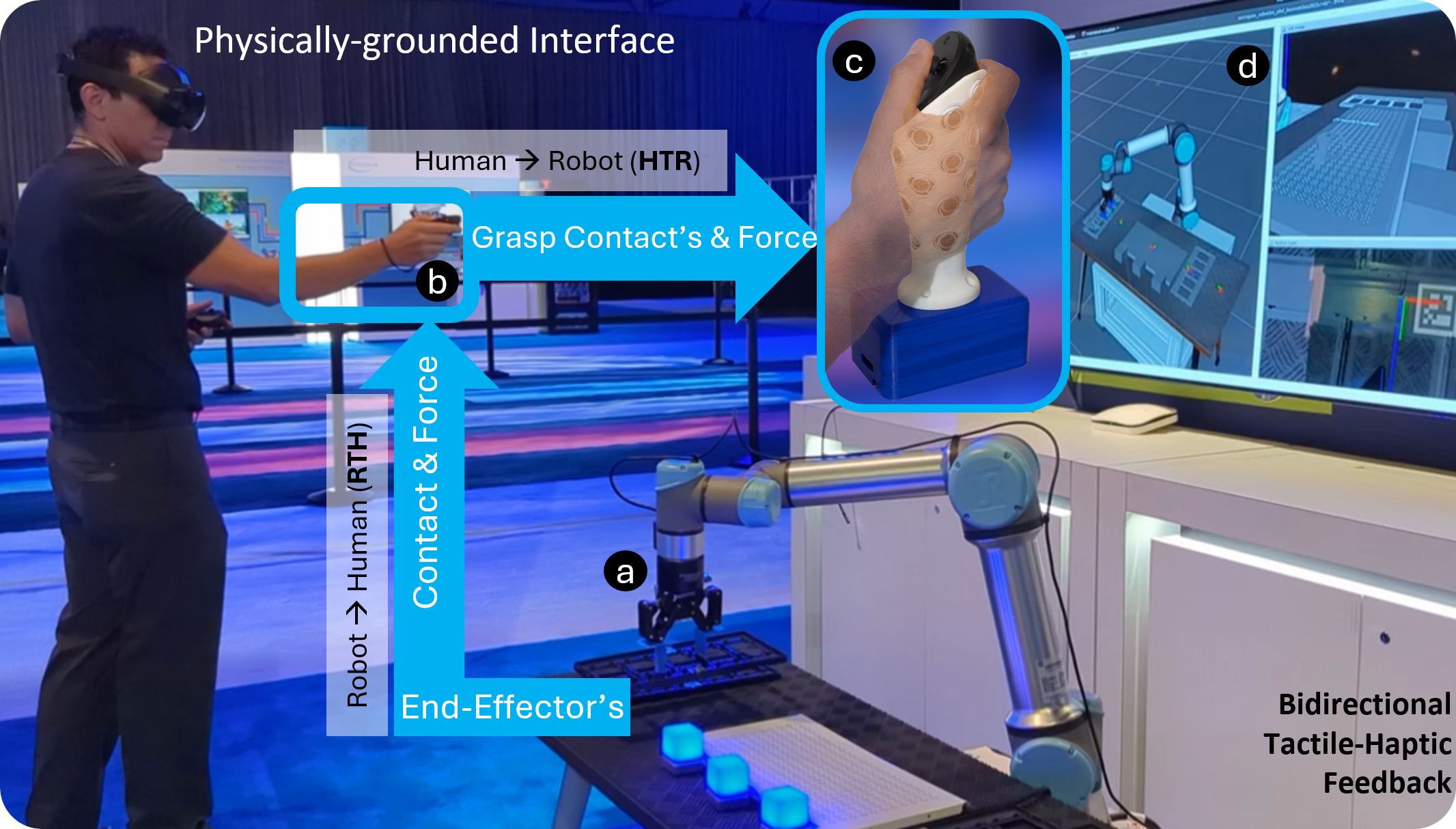}
    \vspace{-3mm}
    \caption{\textbf{Tangible Immersion for Cobot Programming-by-Demonstration.} Cobot programming-by-demonstration through cost-effective and easy-to-deploy mixed reality interfaces are grounded on novel form factors for tactile sensing and haptic feedback interfaces. This approach empowers non-experts to rapidly and intuitively create robot action primitives, composing dependable Cobot task-flows for automation during inspection and manufacturing.}
    \vspace{-4mm}
    \label{fig:Intro_Figure}
\end{figure}
This enables creating robot programs through iconographic flowcharts (see Fig.~\ref{fig:task_representation}), demonstrating Cobots' adaptability in tasks like stacking wafer trays. We aim to minimize programming effort for Cobots, using AI-sensorimotor primitives and hierarchical task graphs. Our goal is to economically democratize robot programmability, fostering opportunities across diverse manufacturing processes.


\section{Methodology}

\subsection{Mixed-Reality for Cobot Programming-by-Demonstration} Creating dependable automation recipes in manufacturing with Cobots requires specialized knowledge of both robotics and the manufacturing process itself. Empowering general users to utilize Cobots to offload tasks is of key value in manufacturing. Consequently, the robotics community has been exploring diverse paradigms to democratize robot programmability. A crucial element in simplifying this programming process is making the state of the environment, robot, tools, and task components observable and manipulable via natural and intuitive means, regardless of their scale, pose, and geographic location. This is where mixed reality provides a convenient set of possibilities by removing the limitations of physical scale and the user's vantage point in the process. Moreover, the need for co-location of user and system is eliminated through a digital twin of the physical world represented in a scene graph, see Fig.\ref{fig:mixed_reality}. Three elements facilitate effective user immersion and agency for programming-by-demonstration:
    \newline\textbf{\textbullet\ Perceptual Immersiveness}: Mixed reality visualizations created from real-time 6D registration using robot perception (section \ref{section:Robot-Action Primitives}) allows users to view the scene from any vantage point and scale, whether standing in the middle of a worktable or making the robot's gripper the size of a room, see Fig \ref{fig:Intro_Figure}-d.
    \newline\textbf{\textbullet\ Spatio-temporal Consistency}: By recognizing objects in the scene using 3D vision and 6D pose estimation methods, either via markers or other mechanisms, the robot setup can register objects in a consistent kinematic tree (via ROS TF). This allows robot and mixed reality processes to share a common state for visualization and contact feedback at any position and scale meeting users and situations needs, see Fig.\ref{fig:mixed_reality}-a,b,c.
    \newline\textbf{\textbullet\ Tangible Actionability}: Integration of head-mounted display and controllers with haptic-feedback by registering to the same kinematic tree allows users to dispatch commands (via ROS topics) from handheld devices at high frame rates, transferring users kinematic frames to transform and mimic human actions in the shared space-time explicitly. Thus, motions, contact actions and associated haptic feedback are grounded in the mixed reality world creating the tangible user experience needed to empower the transfer of factual process knowledge via simple handheld annotated demonstrations, see Fig.\ref{fig:mixed_reality}-d,e,f.
Together, these elements establish a robust framework for translating human intentions and process knowledge into robot actions within a mixed reality environment. Importantly, the physical co-location of the human and robot is optional. Users can register the headset and handheld controllers to a kinematic frame in the physical world, either directly on the robot's support table (Fig.\ref{fig:mixed_reality}-g) or remotely on any surface, ensuring a physical relation when calibrating the system in 6D pose and scale. This decoupling is compatible with both real robot execution and simulation.

\subsection{Robot-Action Primitives}
\label{section:Robot-Action Primitives}
\begin{figure*}
    \centering
    \includegraphics[width=1.0\linewidth]{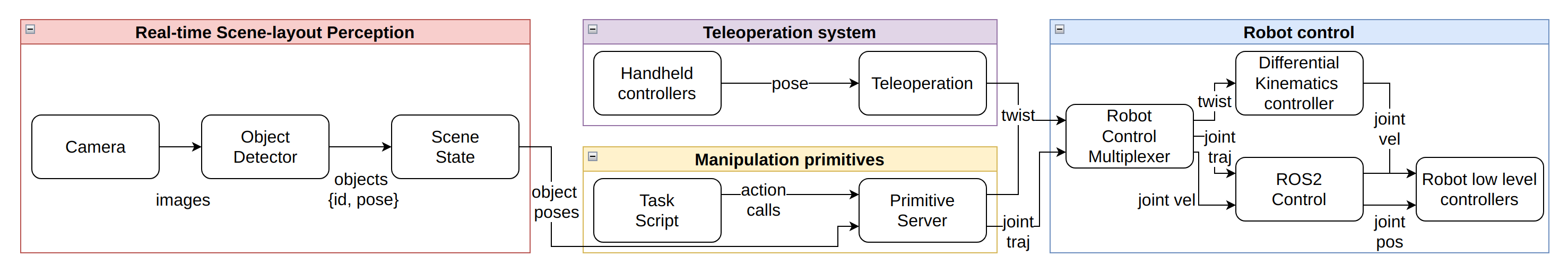}
    \vspace{-9mm}
    \caption{\textbf{Robot-action Primitives System Architecture.}}
    \label{fig:primitives_architecture}
    \vspace{-6mm}
\end{figure*}

The robot action primitives system architecture is designed to facilitate efficient task co-execution via teleoperation as well as autonomous execution by the robot during task offloading. The architecture in  Fig.~\ref{fig:primitives_architecture} consists of four key components:

\subsubsection{Real-time Scene-layout Perception}
The real-time scene-layout perception system supports the robot's ability to interact with its environment. The in-hand RGBD camera continuously monitors the scene, detecting objects and updating their positions using AprilTags \cite{olson2011tags}. While we use AprilTags to detect objects, any other method capable of detecting and localizing objects is suitable to replace and or expand this module.

The scene layout memory system stores information about object 6D pose, allowing the robot to target objects even when they are temporarily out of view. This object permanency feature enables the robot to move towards the last known position of an object, updating its pose as soon as the object re-enters the camera's field of view. Hence, the robot acts on updated 6D poses, ensuring precise and reliable task execution.

\subsubsection{Teleoperation system}
Remote guidance provides a flexible and intuitive method for controlling the robot arm. The operator uses handheld controllers to input pose commands, which are translated into arm movements. The robot replicates the relative movements of the handheld controller, starting from a defined initial position. To enhance operator comfort, the system allows for pausing and resuming teleoperation. 
The robot's movements remain smooth and predictable due to the relative positioning approach, which prevents sudden or erratic motions. This is achieved by calculating an offset transformation for the handheld controller each time teleoperation is activated. The robot is controlled in task space by a PID that computes the twist command $\dot{x}$ based on the difference between the current end-effector pose and the offset-adjusted controller position $e(t) = x_{c}(t)-x(t)$,
\begin{equation}
    \dot{x}(t) = K_p e(t)+ K_i \int_0^t e(t) dt + K_d \frac{de(t)}{dt}. 
	\label{eq:task_space_controller}
\end{equation}
Two sets of gains $\{K_p, K_i, K_d\}$ are employed: a slow (low gain) mode for distant target positions, minimizing sudden movements and signaling a gradual approach, and a fast (high gain) mode for close target positions, enabling quick and precise adjustments for fine and responsive control. Depending on a configurable threshold on $e(t)$ the gains switch between the slow and quick reaction modes.

\subsubsection{Manipulation Primitives}
The robot's ability to perform complex tasks is built upon a set of fundamental action primitives \cite{Felip2013}. These primitives serve as the building blocks for more sophisticated behaviors. Every primitive needs a set of required parameters to be specified and allows for some optional parameters that can fine-tune the behavior of the primitive. See an example of a parameterized grasp primitive in Fig.~\ref{fig:task_representation}, where the \textit{pre-grasp}, \textit{grasp}, \textit{post-grasp}, and other nuanced parameters are defined. The other robot's action primitives include \textit{Move}, \textit{Place}, \textit{LookAt}, and \textit{Perceive}. LookAt orients the camera towards a specific point, while Perceive updates the scene state and identifies objects in view.

\subsubsection{Robot control}
The robot control system integrates several modules. The robot control multiplexer dynamically manages input sources, allowing seamless transitions between teleoperation and autonomous modes. The differential kinematic controller transforms task space twist commands $\dot{x}$ into joint velocities $\dot{\theta}$ using the Jacobian Pseudo-inverse method. For redundant manipulators a secondary joint target $\theta_{sec}$ is projected via nullspace, 
\begin{equation}
    \dot{\theta}(t) = J^{\dagger}(t) \dot{x}(t) + \text{Null}(J^{\dagger}(t)) (\theta_{sec} - \theta(t)).
	\label{eq:joint_space_controller}
\end{equation}
ROS2 control provides a standardized interface for real-time communication and synchronization across components. Low-level controllers directly actuate the robot's joints and motors using velocity or position controllers.

\subsection{Primitive Teaching and Task Description}
The flexible parameterization is what makes the system capable of performing a variety of tasks involving different objects, environments, and action sequences. This is where our mixed reality teleoperation system excels, by enabling users to define how the robot should interact with various sets of objects. Through teleoperation, users specify explicit object-centric parameters that can be encapsulated in object-specific primitives, allowing for tailored interactions based on the unique characteristics of each object. An example is the definition of the grasp primitive (see Fig.~\ref{fig:mixed_reality}-d,e,f) which involves teleoperating the robot's end effector to the pre-grasp, grasp, and post-grasp poses deemed appropriate for the target object category. This approach allows users to impart their expertise in handling tasks without any robotics knowledge.

\subsection{Task descriptions and autonomous execution}
\begin{figure*}[t]
    \centering
    \includegraphics[width=1.0\linewidth]{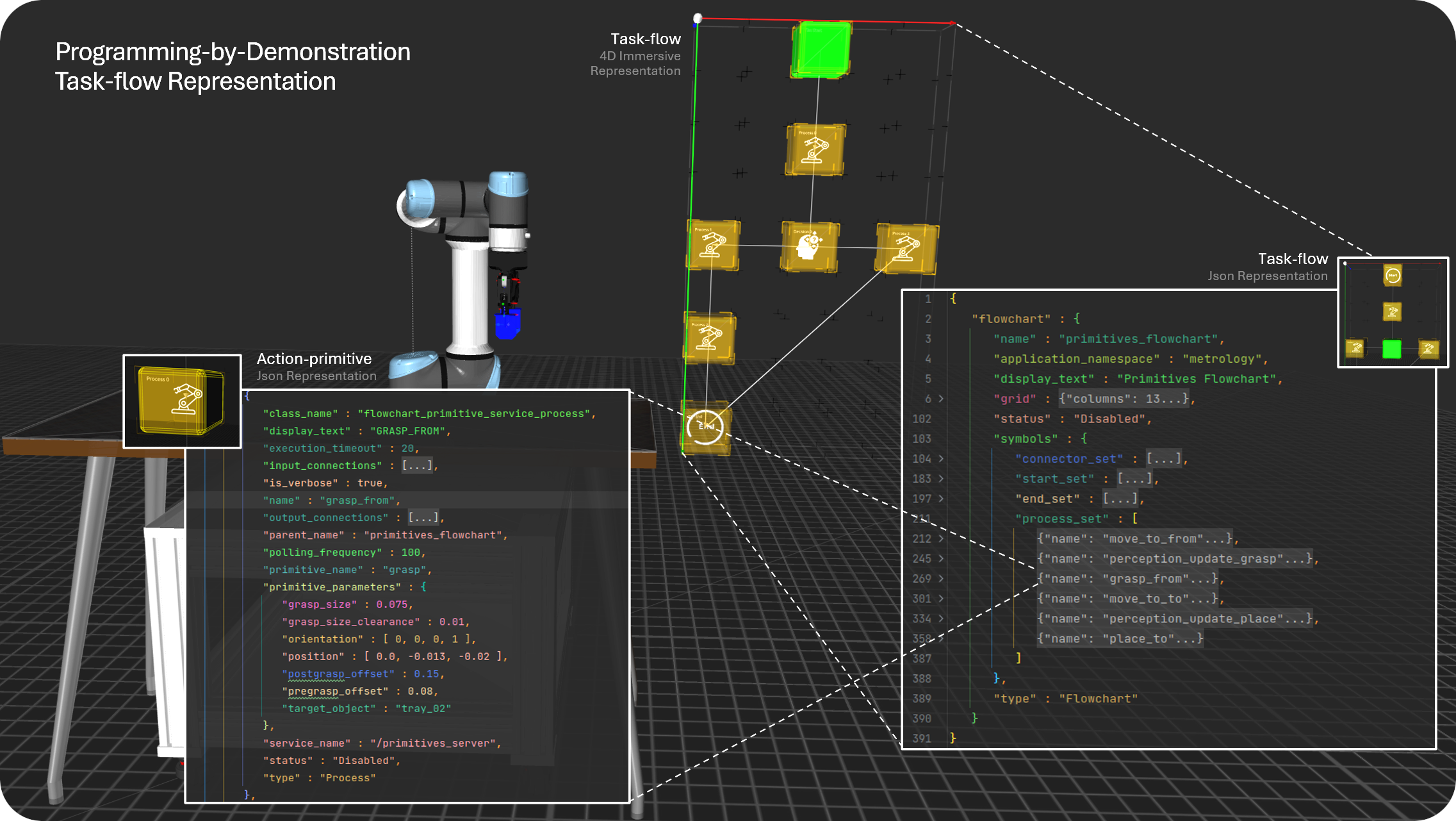}
    \vspace{-7mm}
    \caption{\textbf{Robot Task Representation via Action Primitives}. A sequence of modular primitives are combined to execute complex tasks autonomously. Each primitive is grounded on perceived objects, allowing for flexible and adaptable task execution. The zoomed-in views display the JSON files generated which describe the task sequence and the parameters for each primitive, highlighting the structured and programmable nature of task configuration.}
    \label{fig:task_representation}
\end{figure*}

When the individual primitives are parameterized for each of the objects, they can be composed to perform complex tasks. The perceived action is often the initial step in a sequence of primitives, providing the necessary context for subsequent actions such as move, grasp, and place. These primitives are designed to be modular and adaptable, enabling the robot to perform a wide range of tasks by combining and sequencing them as needed. By leveraging these fundamental actions, the robot can learn and execute tasks through simple scripts that sequence actions grounded on perceived objects. An example of this representation is shown in Fig.~\ref{fig:task_representation}, illustrating how tasks are structured and executed autonomously.

\begin{figure*}
    \centering
    \vspace{-0mm}
    \includegraphics[width=1\linewidth]{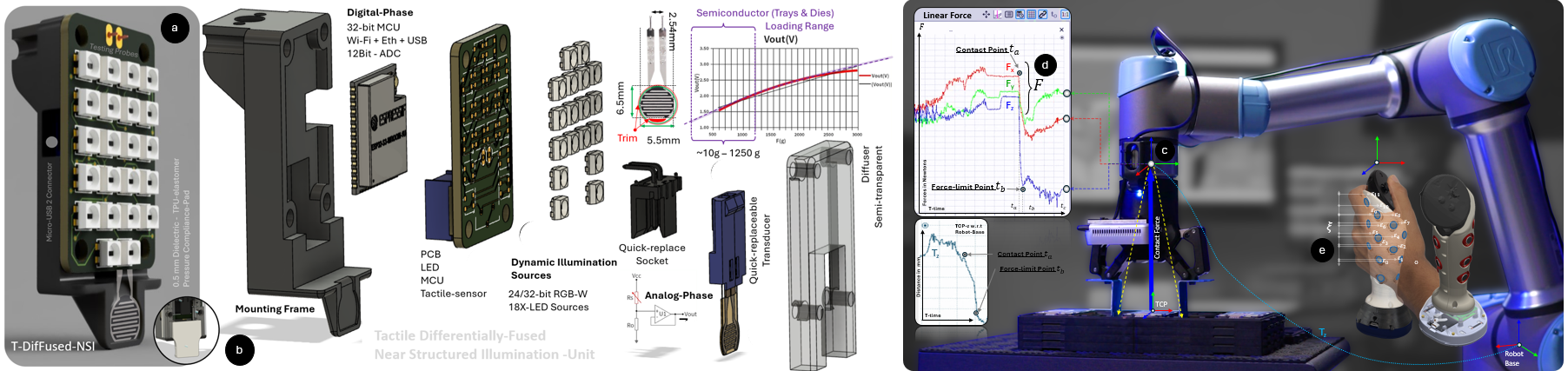}
    \caption{\textbf{Novel Tactile Sensing and Haptic Feedback Devices.} a) The integrated active illumination and b) replaceable-transducer in the tactile sensor capture contact's pressure at high frequency (1 Khz) without contamination materials for easy and economic deployment. The haptic handheld feedback device allows the user to sense c) contacts and forces applied to and by the robot with d) low-latency ($\sim$2 ms) and low-cognitive load e) opening new channels of effective communication between humans and robot in mixed reality.}
    \label{fig:novel_devices}
    \vspace{-5mm}
\end{figure*}
\subsection{Tactile Sensing}
Handling small, fragile objects in semiconductor foundries and high-precision applications presents significant challenges due to the absence of delicate, pressure-aware handling solutions. Current sensors often lack the necessary sensitivity, dynamic range for both lighting and pressure detection, and high-resolution capabilities required for these tasks. Additionally, existing sensing technologies, such as flexible polymers, resistive foams, or metal/polymer coil arrays, are unsuitable for opto-electronic interfaces and pharma-biotechnological processes \cite{goncalves2022punyo1softtactilesensingupperbody, lambeta2024digitizingtouchartificialmultimodal}. Frequent offline re-calibrations of commercial sensors further exacerbate manufacturing time and costs. 

Current tactile solutions lack the necessary form factors and integration, limiting their effectiveness \cite{tactile_survey},\cite{al_2020_tactile_survey}. To address these limitations, we propose a solution that leverages collaborative robots equipped with innovative sensor-actor units and visuo-haptic algorithms, see Fig.\ref{fig:novel_devices}-a. These units are designed to achieve beyond-human precision in delicate object handling, offering automated material handling that reduces risks associated with cleanliness, damage, and human error. 

By fusing real-time information, this approach dynamically avoids obstacles and optimizes pickup position, orientation, and lighting for each situation, accommodating sample sizes from 7 to 200 millimeters, see Fig.\ref{fig:novel_devices}-b. This visuo-haptic approach provides a single, cost-effective solution that maintains exceptional quality and cleanliness compared to human operators, enhancing process quality, yield, and throughput in high-mix, low-volume situations with delicate parts.

\subsection{Haptic Feedback}

\subsubsection{Motivation and Limitations of State-of-the-art}

The motivation for enhancing haptic feedback in mixed reality systems is driven by the need to improve user immersion and interaction precision, particularly in teleoperation tasks. Current mixed reality controllers primarily offer low-dimensional vibrotactile feedback, resulting in monolithic vibrations that fail to convey detailed tactile information necessary for complex tasks \cite{PA_Cabaret}. While advancements have been made in motion tracking and ergonomics, the haptic capabilities of these devices remain limited. Our prototype (see Fig\ref{fig:novel_devices}-e) addresses these limitations by integrating a flexible membrane with strategically placed linear resonant actuators (LRAs), providing localized and varied tactile sensations. This design enhances the richness of haptic feedback, crucial for improving user interaction precision and immersion without disrupting the teloperation experience, as compared to kinesthetic force feedback that is prone to have a harmful effect on the stability of the control loop \cite{Pacchierrotti}.

\subsubsection{Design Principles and Key Features}

The design of our haptic prototype focuses on maximizing user immersion and interaction precision through a membrane with strategically placed LRAs. The actuators are aligned with the receptivity zones of the human hand, particularly at the distal parts of the fingers where mechanoreceptors are densely packed \cite{Vallbo1984Mechanoreceptors}. This placement ensures effective stimulation of both Rapid Adapting (RA) and Slow Adapting (SA) mechanoreceptors, providing rich tactile cues for motion guidance and force perception \cite{DING2016131}. The high-density array of LRAs allows for the conveyance of multi-dimensional haptic cues and patterns. Figure \ref{fig:novel_devices}-e illustrates a use case scenario in which the tactile array is used to render intuitive haptic cues that indicate both the direction and intensity of contact forces perceived at the robot's end effector. These features enhance the overall telepresence experience for the operator, making interactions more natural and immersive.

\subsubsection{Contact and Contactless State Disclosure with Low-latency and Low-cognitive load}

Our prototype excels in disclosing contact and contactless states with minimal latency and cognitive load, enhancing user immersion and interaction precision. To achieve this, the device utilizes embedded haptic patterns stored in its memory, allowing for the rendering of vibrations based on local information rather than relying on the continuous transmission of large amounts of data. This approach significantly reduces latency, ensuring rapid and responsive feedback. The device features a soft mechanism to isolate mechanical LRA waves across the membrane, ensuring localized stimulation across different hand regions. This design supports low cognitive load interactions by allowing the user to discriminate tactile stimulation, see Fig.\ref{fig:novel_devices}-d. Enhancing the haptic capabilities of mixed reality handheld controllers significantly improves user situational awareness and spatial coordination during teleoperation tasks, making the prototype an optimal solution for high-mix, low-volume teleoperation in manufacturing and other applications.

\section{Discussion and Conclusion}
We introduced and integrated three innovative technology components: 1) programming-by-demonstration using mixed reality for local and remote teleoperation and task automation, 2) advanced form factors for sensing and haptic feedback that establish a tangible connection with the physical world during the creation of robot action primitives, and 3) a task description framework based on modular and adaptable robot action primitives for synthesizing complex tasks.

These technologies collectively form a unique framework that offers several benefits: (1) high precision in task execution, achieving millimeter accuracy, (2) an intuitive immersive user interface, ensuring accurate perception of the digital twin and tangible sensing of interactions, and (3) efficient creation of robot tasks by demonstrating a sequence of actions within a real scene. This framework not only enhances the capabilities of collaborative robots but also democratizes their programmability, making advanced automation accessible to a broader range of users and applications across industries.



\bibliographystyle{IEEEtran}
\bibliography{bibliography.bib}

\begin{thebibliography}{10}
\providecommand{\url}[1]{#1}
\csname url@samestyle\endcsname
\providecommand{\newblock}{\relax}
\providecommand{\bibinfo}[2]{#2}
\providecommand{\BIBentrySTDinterwordspacing}{\spaceskip=0pt\relax}
\providecommand{\BIBentryALTinterwordstretchfactor}{4}
\providecommand{\BIBentryALTinterwordspacing}{\spaceskip=\fontdimen2\font plus
\BIBentryALTinterwordstretchfactor\fontdimen3\font minus \fontdimen4\font\relax}
\providecommand{\BIBforeignlanguage}[2]{{%
\expandafter\ifx\csname l@#1\endcsname\relax
\typeout{** WARNING: IEEEtran.bst: No hyphenation pattern has been}%
\typeout{** loaded for the language `#1'. Using the pattern for}%
\typeout{** the default language instead.}%
\else
\language=\csname l@#1\endcsname
\fi
#2}}
\providecommand{\BIBdecl}{\relax}
\BIBdecl

\bibitem{stepputtis2020language}
S.~Stepputtis, J.~Campbell, M.~Phielipp, S.~Lee, C.~Baral, and H.~Ben~Amor, ``Language-conditioned imitation learning for robot manipulation tasks,'' \emph{Advances in Neural Information Processing Systems}, vol.~33, pp. 13\,139--13\,150, 2020.

\bibitem{black2024pi_0}
K.~Black, N.~Brown, D.~Driess, A.~Esmail, M.~Equi, C.~Finn, N.~Fusai, L.~Groom, K.~Hausman, B.~Ichter \emph{et~al.}, ``$\\\pi_0$: A vision-language-action flow model for general robot control,'' \emph{arXiv preprint arXiv:2410.24164}, 2024.

\bibitem{hu2023robofm}
Y.~Hu, Q.~Xie, V.~Jain, J.~Francis, J.~Patrikar, N.~Keetha, S.~Kim, Y.~Xie, T.~Zhang, H.-S. Fang, S.~Zhao, S.~Omidshafiei, D.-K. Kim, A.~akbar Agha-mohammadi, K.~Sycara, M.~Johnson-Roberson, D.~Batra, X.~Wang, S.~Scherer, C.~Wang, Z.~Kira, F.~Xia, and Y.~Bisk, ``Toward general-purpose robots via foundation models: A survey and meta-analysis,'' \emph{arXiv preprint: arXiv:2312.08782}, 2023.

\bibitem{olson2011tags}
E.~Olson, ``{AprilTag}: A robust and flexible visual fiducial system,'' in \emph{Proceedings of the {IEEE} International Conference on Robotics and Automation ({ICRA})}.\hskip 1em plus 0.5em minus 0.4em\relax IEEE, May 2011, pp. 3400--3407.

\bibitem{Felip2013}
\BIBentryALTinterwordspacing
J.~Felip, J.~Laaksonen, A.~Morales, and V.~Kyrki, ``Manipulation primitives: {A} paradigm for abstraction and execution of grasping and manipulation tasks,'' \emph{Robotics and Autonomous Systems}, vol.~61, no.~3, pp. 283--296, Mar. 2013. [Online]. Available: \url{http://www.sciencedirect.com/science/article/pii/S0921889012002217}
\BIBentrySTDinterwordspacing

\bibitem{goncalves2022punyo1softtactilesensingupperbody}
\BIBentryALTinterwordspacing
A.~Goncalves, N.~Kuppuswamy, A.~Beaulieu, A.~Uttamchandani, K.~M. Tsui, and A.~Alspach, ``Punyo-1: Soft tactile-sensing upper-body robot for large object manipulation and physical human interaction,'' 2022. [Online]. Available: \url{https://arxiv.org/abs/2111.09354}
\BIBentrySTDinterwordspacing

\bibitem{lambeta2024digitizingtouchartificialmultimodal}
\BIBentryALTinterwordspacing
M.~Lambeta, T.~Wu, A.~Sengul, V.~R. Most, N.~Black, K.~Sawyer, R.~Mercado, H.~Qi, A.~Sohn, B.~Taylor, N.~Tydingco, G.~Kammerer, D.~Stroud, J.~Khatha, K.~Jenkins, K.~Most, N.~Stein, R.~Chavira, T.~Craven-Bartle, E.~Sanchez, Y.~Ding, J.~Malik, and R.~Calandra, ``Digitizing touch with an artificial multimodal fingertip,'' 2024. [Online]. Available: \url{https://arxiv.org/abs/2411.02479}
\BIBentrySTDinterwordspacing

\bibitem{tactile_survey}
\BIBentryALTinterwordspacing
W.~Mandil, V.~Rajendran, K.~Nazari, and A.~Ghalamzan-Esfahani, ``Tactile-sensing technologies: Trends, challenges and outlook in agri-food manipulation,'' \emph{Sensors}, vol.~23, no.~17, 2023. [Online]. Available: \url{https://www.mdpi.com/1424-8220/23/17/7362}
\BIBentrySTDinterwordspacing

\bibitem{al_2020_tactile_survey}
Y.~Al-Handarish, O.~M. Omisore, T.~Igbe, S.~Han, H.~Li, W.~Du, J.~Zhang, and L.~Wang, ``A survey of tactile-sensing systems and their applications in biomedical engineering,'' \emph{Advances in Materials Science and Engineering}, vol. 2020, no.~1, p. 4047937, 2020.

\bibitem{PA_Cabaret}
\BIBentryALTinterwordspacing
P.-A. Cabaret, T.~Howard, C.~Pacchierotti, M.~Babel, and M.~Marchal, ``Perception of spatialized vibrotactile impacts in a hand-held tangible for virtual reality,'' in \emph{Haptics: Science, Technology, Applications: 13th International Conference on Human Haptic Sensing and Touch Enabled Computer Applications, EuroHaptics 2022, Hamburg, Germany, May 22–25, 2022, Proceedings}.\hskip 1em plus 0.5em minus 0.4em\relax Berlin, Heidelberg: Springer-Verlag, 2022, p. 264–273. [Online]. Available: \url{https://doi.org/10.1007/978-3-031-06249-0\_30}
\BIBentrySTDinterwordspacing

\bibitem{Pacchierrotti}
C.~Pacchierotti and D.~Prattichizzo, ``Cutaneous/tactile haptic feedback in robotic teleoperation: Motivation, survey, and perspectives,'' \emph{IEEE Transactions on Robotics}, vol.~40, pp. 978--998, 2024.

\bibitem{Vallbo1984Mechanoreceptors}
A.~B. Vallbo and R.~S. Johansson, ``Tactile sensory coding in the glabrous skin of the human hand,'' \emph{Trends in Neurosciences}, vol.~7, no.~2, pp. 27--32, 1984.

\bibitem{DING2016131}
\BIBentryALTinterwordspacing
S.~Ding and B.~Bhushan, ``Tactile perception of skin and skin cream by friction induced vibrations,'' \emph{Journal of Colloid and Interface Science}, vol. 481, pp. 131--143, 2016. [Online]. Available: \url{https://www.sciencedirect.com/science/article/pii/S0021979716304933}
\BIBentrySTDinterwordspacing

\end{thebibliography}

\end{document}